\documentclass[eng]{ajceam-class}
\usepackage{graphicx}
\usepackage{filecontents}

\title{Handling and extracting key entities from customer conversations using Speech recognition and Named Entity recognition }

\author{Sharvi Endait}
\author{Ruturaj Ghatage}
\author{Prof. D.D Kadam}

\affil[1]{Pune Institute of Computer Technology, Pune, India}
\firstauthor{}

\abstract{
In this modern era of technology with e-commerce developing at a rapid pace, it is very important to understand customer requirements and details from a business conversation. It is very crucial for customer retention and satisfaction. Extracting key insights from these conversations is very important when it comes to developing their product or solving their issue. Understanding customer feedback, responses, and important details of the product are essential and it would be done using Named entity recognition (NER). For extracting the entities we would be converting the conversations to text using the optimal speech-to-text model. The model would be a two-stage network in which the conversation is converted to text. Then, suitable entities are extracted using robust techniques using a NER BERT transformer model.
This will aid in the enrichment of customer experience when there is an issue which is faced by them. If a customer faces a problem he will call and register his complaint. The model will then extract the key features from this conversation which will be necessary to look into the problem. These features would include details like the order number, and the exact problem. All these would be extracted directly from the conversation and this would reduce the effort of going through the conversation again.}


\begin{document}

\maketitle
\thispagestyle{fancy}

\section{Introduction}
\firstword{T}{  he}
Research in natural language processing aims to gather knowledge on how human beings understand and use a particular language so that appropriate tools and techniques can be developed to make computer systems understand languages to perform the desired task. NLP is the ability of a computer program to understand human language as it is spoken and written.  \\

ASR( Automatic Speech Recognition) deals with audio inputs and delivers a transcript on a particular device. Data coming from customer care service phone calls/products or a service inquiry call/ product or a service complaint call or a service review call recordings can come through by passing them through a speech-to-text converter. Sensing vibrations via the phone call recordings and measuring and detailing those waves to distinguish the relevant sounds land in a suitable format to segment the data to match them with phenomes that are run through a mathematical model. Fairly accurate textual data is obtained through this process which can save the conversation transcription for further processing.

The conversation audio is passed to the model which is based on Wav2Vec transformer. It is a state of the art model which is trained in 2 phases. It is an Automatic Speech Recognition model based on self supervised training.[12][15] The two phases include:\\

    1. Self supervised Mode: This contains the unlabelled data.\\
    
    2. Supervised Fine tuning: during which the labelled data is used to teach model to predict particular words or phenomes.

For extracting primary details such as name, residential area, product details such as the brand, its pricing, and detailing regarding that product, Named Entity Recognition can be used for processing that data. It involves the identification of key information in the text and classification into a set of predefined categories. Here, an entity is a key that is consistently referred to in the text.

NER involves detecting the entities initially and then classifying the data into different categories. Here in this case although the classification of entities might not play a major role, the detection of the entities accurately and including all key details seem to be a necessity. NER can be done using multi-class classification, NLP speech tagging, and by using Deep learning. The model that is aimed to be used is the BERT-base-NER which is a fine-tuned Bidirectional Encoder Representation from Transformers. It's primarily trained to recognize location, organizations, persons, and miscellaneous details.

 Deep learning can be categorized into various domains, among which transformers are a fairly recent development in the field. Transformers are designed to process sequential data and this process is done in one go. The context to every word is understood from the position in the input sequence. Transformer is an architecture for transforming one sequence to another with the help of 2 parts: Encoder and Decoder.

\includegraphics[scale=0.5]{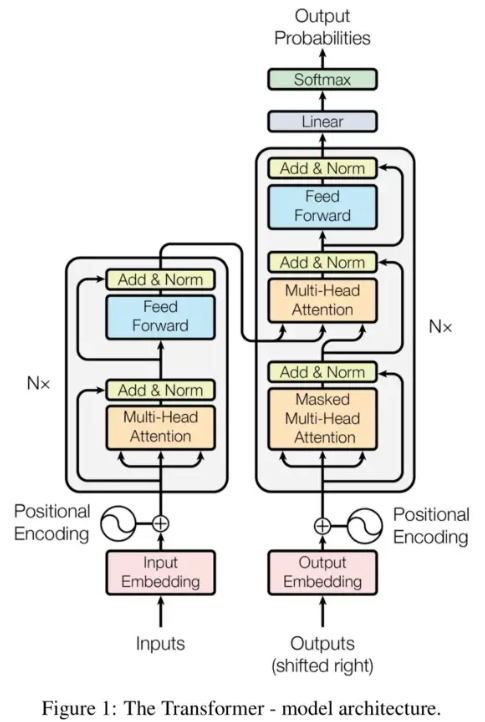}
\textit{Image Courtesy :The
credit of this image is for (Vaswani et al., 2017). }\\
The Encoder is on the left and the Decoder is on the right. Both Encoder and Decoder are composed of modules that can be stacked on top of each other multiple times.

\section{Motivation}

As technology has progressed, the amount of data generated has grown exponentially. The business sector needs an efficient way to manage and handle data. So there has to be a productive way you can systematically gather the information and specific entities from the clients for the betterment and upgradation of their businesses. Customer reviews and customer satisfaction is a key factor in the improvisation of the product/service offered by the company. 

As machine learning and artificial intelligence have been developing rapidly since the dawn of this century, various new algorithms and technologies are coming forward which are used to tackle a vast set of problems. Natural Language Processing is a subset of AI/ML that deals with giving computers the ability to understand text and spoken words in much the same way human beings can. Handling customer reviews can be simplified by using suitable NLP algorithms. 

\section{Scope}

To understand the customer details transcribing the audio from the calls recorded using speech recognition Wav2Vec model, and then extracting key entities from the text generated using Named Entity Recognition. This simplifies the entire process of understanding the customer's review and would help the company for the betterment of the product/services.

\section{Literature Survey}

\subsection{Speech Recognition models:}

\subsubsection{Deep Speech by Baidu and Listen Attend Spell (LAS) by Google:}

Deep Speech and Listen Attend Spell both use recurrent neural network-based architectures for speech recognition. Deep Speech uses the Connectionist Temporal Classification (CTC) loss function to predict the speech transcript. LAS uses a sequence-to-sequence network architecture for its predictions. These models are built and trained in such a way that they can build and understand from larger datasets. With enough data, in theory, one would be able to build a super robust speech recognition model that can account for all the nuance in speech without having to spend a ton of time and effort hand-engineering acoustic features or dealing with complex pipelines in more old-school GMM-HMM model architectures.[3]

\subsubsection{Wav2Vec Model:}

Wav2vec is a speech recognition algorithm that uses raw, unlabeled audio to train automatic speech recognition (ASR) models which beats the traditional ASR systems that rely solely on transcribed audio, including a 22 percent accuracy improvement over Deep Speech 2 while using two orders of magnitude less labeled data.Wav2vec trains models to learn the difference between original speech examples and modified versions, often repeating this task hundreds of times for each second of audio and predicting the correct audio milliseconds into the future. Wav2Vec reduces the need for data to be manually annotated for developing systems of limited training sets and non english languages. Wav2Vec 2.0 comes under the category of self-supervised learning and development which requires self-training, where a teacher model is used to improve the performance of a student model by generating labels for the unlabeled set the student can train on. Self-training has been a popular technique in ASR studied extensively in the literature [7][15].\\

There is little domain mismatch between the unlabeled data for pre-training, the labeled data for fine-tuning and the domain of the test data, or the target domain. Domain is like an area where we are particularly working. Due to this, the performance of ASR systems trained from scratch with conventional supervised objectives can degrade significantly when tested on domains mismatched from training data [5]. Adding unlabeled in-domain data improves performance, even when the fine-tuning data does not match the
test domain[6]. The need for large annotated data in training remains a challenge for research in this area, especially for specific domains as per survey [8].

\begin{figure*}[thpb]
    \centering
    \includegraphics[width=\textwidth, height=110mm]{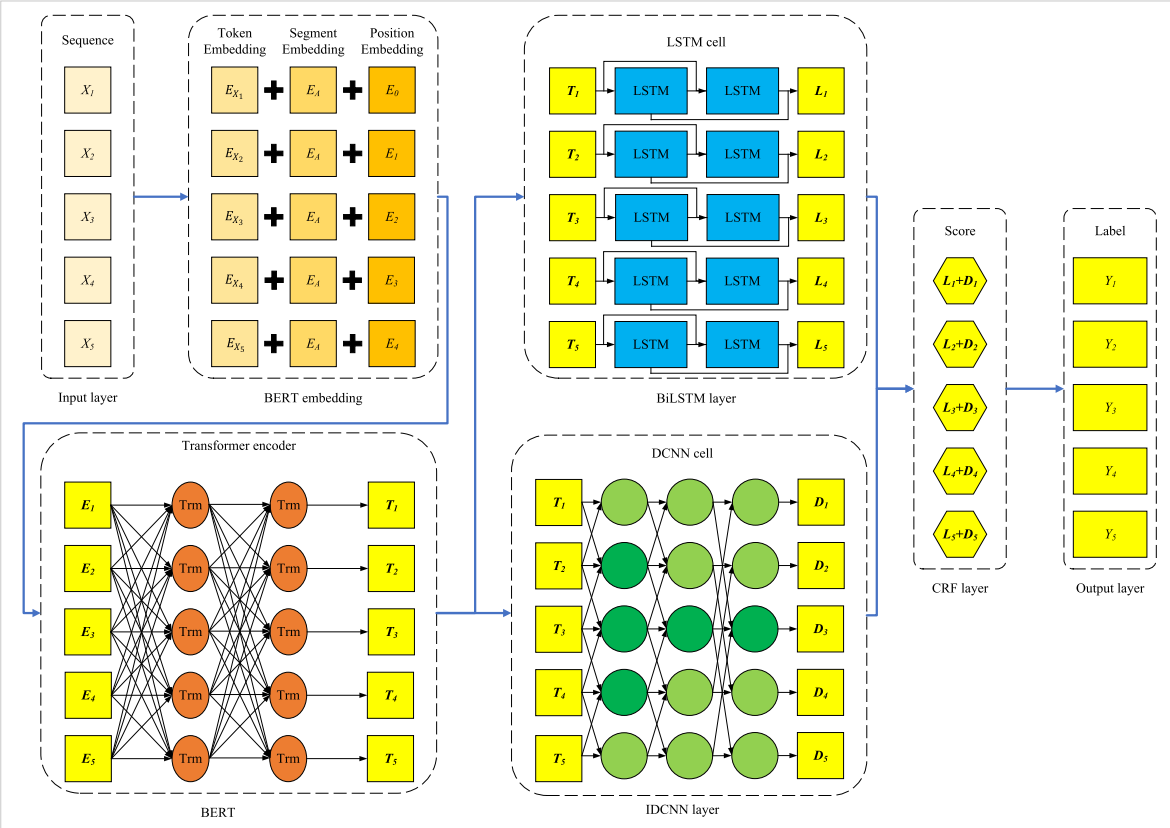}
    \caption{NER based on BERT}\vspace{4 mm}
     \textit{Image Courtesy: S. Hu, H. Zhang, X. Hu and J. Du, "Chinese Named Entity Recognition based on BERT-CRF Model,"  }
    \label{fig:my_label}
\end{figure*}

\subsection{Named entity recognition models}
\subsubsection{Ontology-based NER}
Ontology-based NER refers to understanding through knowledge bases that are normally a collection of datasets containing terms and their interrelations. This type of learning strongly relies on updates. Otherwise, it can’t keep up with the ever-growing publicly available knowledge. This model includes a knowledge-based recognition process where datasets containing terms are interrelated together using dictionaries and other mapping techniques.[2]

\subsubsection{Deep learning NER}
Deep Learning NER is much more precise than its Ontological NER  as it is able to cluster words. This could be possible due to word embedding, that can understand the syntactic and semantic relationship between words. Deep learning can recognize terms \& concepts not present in Ontology because it is trained on the way various concepts are used in the written life science language. It is able to learn automatically and analyzes topic-specific as well as high-level words. This makes deep learning NER applicable for a variety of tasks. Researchers, for example, can use their time more efficiently as deep learning does most of the repetitive work. They can focus more on research. Currently, there are several deep-learning methods for NER available. But due to competitiveness and recency of developments, it is difficult to pinpoint the best one on the market. [4]

A machine learning technique is also presented in [13] to improve the precision of extracting entities related to organizations and persons. The first seed utilizes building and training the machine learning algorithm for some categories across domains, languages, and other applications. Multiple NER models can be constructed and implemented in order to emphasize the importance of certain architectural elements of the model as well as the features being used. [11]

\subsubsection{NER based on BERT}
The bert-based NER is divided into 3 layers:
the BERT pre-training layer, BiLSTM-IDCN-ELU neural network layer, and the CRF inference layer	

\emph{Pre-training:} in this step, the corpus is divided into 2 levels: word level and character level. the effect of entity recognition based on character vectors is more accurate. the BERT embedding layer is chosen to add position information to the characters, and its structure consists of Token Embeddings, Segment Embeddings, and Position Embeddings, respectively.[1]

\begin{figure*}
    \centering
    \includegraphics[width=\textwidth]{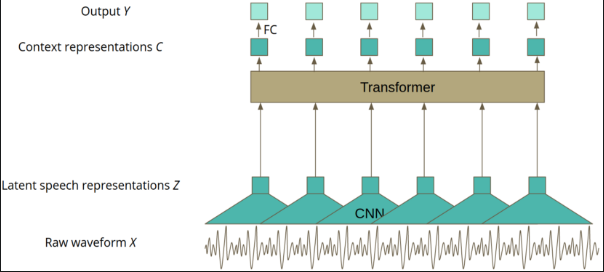}
    \caption{Wav2Vec Model}\vspace{4 mm}
    \textit{Image Courtesy: https://neurosys.com/blog/wav2vec-2-0-framework}
\end{figure*}

\emph{BiLSTM-IDCNN-ELU Neural Network Layers:} Long Short-Term Memory Neural Networks are a special kind of RNN. Its neurons have three parts: input gate, forgetting gate, and output gate. The input gate controls which information is input, the forgetting gate controls which information is forgotten in the neuron, and the output gate controls which information is output. to avoid the problem of losing the historical features due to the long sentences by splicing the features in both directions, to better learn the contextual features and solve the long-distance dependency problem so we use bidirectional LSTM.
Dilated Convolutional Neural Networks (DCNN) are a particular type of CNN with a convolution kernel that adds a dilation distance d, which can learn local features better. Iterated Dilated Convolutional Neural Networks (CNN) are composed of multiple layers of DCNNs with different dilation widths.ELU is Exponential Linear Unit. [1]\\

\emph{CRF:} They can effectively check the labels of columns and improve recognition accuracy. Therefore, CRF Conditional random fields are used as an inference layer to avoid label position errors.[1] Deep learning has developed rapidly and
received increasingly attention, and neural network meth-
ods are widely used in many NLP tasks, including the
CNER task. Neural network approaches, especially for
BiLSTM-CRF model, can significantly
improve the performance of the NER task.[9][10]

\section{Model Description}

\subsection{Speech Recognition:}

Raw data is more easily available and accessible. This raw data is not labelled. self-supervised learning is such a new technique which is beneficial in pre-training a model which is unlabelled. Wav2Vec 2.0 is a state of the art model for Automatic Speech Recognition using this self supervised training.



The Wav2Vec model is trained in 2 phases. The first phase is the pre-training phase, which used the unlabelled data, does self-supervised learning on it and attempts to achieve best speech representation as possible. This basically generates word embeddings of audio. The second phase of training is supervised fine tuning, here a labelled data is used to teach the model to predict particular words or phenomes.[14] Phenome here refers to a unit of sound. For this second phase, an appropriate labelled training set is a requirement which is not easily available. This phase tailors the model to predict words according to our own requirement.\vspace{0.8mm}

Wav2Vec 2.0 Model Architecture
The architecture of the final model used for prediction is divided into 3 main parts:
convolutional layers that process the raw waveform input to get latent representation - Z,
transformer layers, creating contextualized representation - C,
linear projection to output - Y.\vspace{0.8mm}

Contrastive learning is a machine learning technique in which unlabelled data points are placed against each other to determine their similarities and their differences. Self-supervised training is based on the idea of contrastive learning. As labelled dataset is not readily available, this technique is of immense help in getting satisfying results.\vspace{0.8mm}


It uses convolutional layers to preprocess raw waveform and then it applies transformer to it. Transformer enhances the speech representation with context.


\subsection{Named Entity Recognition}

BERT stands for Bidirectional Encoder Representations from Transformers. This is a pre-trained transformer. This model was trained based keeping in mind the left and right contexts of a particular word. As both the right and left contexts are considered, all words in the sentence are considered. Their effect on the focus word is considered.

BERT model achieves state-of-the-art accuracy on several tasks as compared to other RNN architectures. However, it needs high computational power and it takes a lot of time to train a model. After this we fine tune the model to get our desired results. 

BERT is a language transformer heavily based on Transformers. It takes embedding tokens of one or more sentences as input. The first token is called as CLS. The sentences are separated by tokens called SEP. BERT outputs an embedding token called hidden state. Bert was trained on the masked language model and next sentence prediction tasks.

In the masked language model (MLM), an input word (or token) is masked and BERT has to try to figure out what the masked word is. The next task is next sentence prediction (NSP), here 2 sentences are taken as input. BERT then figures out whether the second sentence sematically follows the first sentence or not.

If you think about it, solving the named entity recognition task means classifying each token with a label like person name, location, etc. The hidden state of each token is obtained and fed to a classification layer.

The words of a sentence are tokenized. In this model, we have used bert-base-NER. 

BERT uses a method of masked language modeling to keep the word in focus from "seeing itself" or having a fixed meaning independent of its context. BERT then identifies the masked word based on context alone. In BERT, words are defined by their surroundings. It relies on self attention mechanism.



\section{Algorithm}

\begin{enumerate}
    \item The recorded audio is passed to the speech recognition model. The speech recognition model is based on transformers, particularly Wav2Vec.
    \item The model transcribes the speech to text and gives it as an output.
    \item The generated output is then based on the Entity recognition model which is based on the BERT transformer.
    \item The model then extracts the entities from the generated text and presents them as the output of the process.
    \item The output contains the important information which is present in the audio like the order id and thus makes it easier to get information easily from conversations.
\end{enumerate}

\section{Limitations and Future work}

\subsection{Limitations}
This research provides the first step toward understanding customer requirements and reviews through conversations. The main limitation is the efficiency of the model due to the unavailability of sufficient data regarding this domain and the specific design of a platform that can handle this scenario.

The main problem here lies during NER, as the model is predefined we just need to fine tune it. This requires a properly annotated dataset. The topic of our choosing is extracting entities from customer conversations, so there should be a dataset available for the same which isnt readily available. So the requirement of the dataset is a limitation for the model's fine-tuning.

\subsection{Future Work}
Design of an end-to-end application or development of an API that can directly extract entities with a highly accurate recognition model.

\begin{center}
    \includegraphics[scale=0.6]{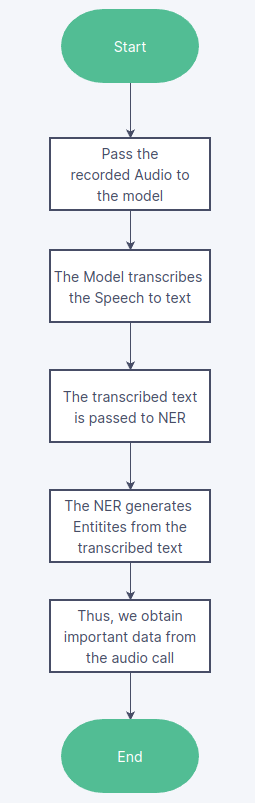}\\
    \textbf{Fig 4.} Flowchart of working
\end{center}
\section{Conclusions}

Using the methodology of Transfer learning we have fine-tuned the model to extract appropriate entities from the customer conversations that would help in the improvement of the business and services. With the help of Named Entity Recognition models and Speech recognition models, customer reviews can be efficiently extracted and that can work in the enhancement of the performance of the product. Wav2Vec model proved to be a suitable model when it comes to the efficient speech to text translation. \\

\bibliographystyle{plainnat}
\bibliography{Bibliografy}

[1] S. Hu, H. Zhang, X. Hu and J. Du, "Chinese Named Entity Recognition based on BERT-CRF Model," 2022 IEEE/ACIS 22nd International Conference on Computer and Information Science (ICIS), 2022, pp. 105-108, doi: 10.1109/ICIS54925.2022.9882514.\vspace{1 mm}

[2] O. Çiftçi and F. Soygazi, "Ontology Based Turkish Named Entity Recognition," 2020 Innovations in Intelligent Systems and Applications Conference (ASYU), 2020, pp. 1-5, doi: 10.1109/ASYU50717.2020.9259897.\vspace{1 mm}

[3] Aashish Agarwal and Torsten Zesch, "German End-to-end Speech Recognition based on DeepSpeech", Proceedings of the 15th Conference on Natural Language Processing (KONVENS 2019), 2019\vspace{1 mm}

[4] S. Anbukkarasi, S. Varadhaganapathy, S. Jeevapriya, A. Kaaviyaa, T. Lawvanyapriya and S. Monisha, "Named Entity Recognition for Tamil text Using Deep Learning," 2022 International Conference on Computer Communication and Informatics (ICCCI), 2022, pp. 1-5, doi: 10.1109/ICCCI54379.2022.9740745.\vspace{1 mm}

[5] Tatiana Likhomanenko, Qiantong Xu, Vineel Pratap, Paden Tomasello, Jacob Kahn, Gilad Avidov, Ronan
Collobert, and Gabriel Synnaeve. Rethinking evaluation in asr: Are our models robust enough? arXiv, 2020.\vspace{1 mm}

[6] Wei-Ning Hsu, Anuroop Sriram and Alexei Baevski et al. Robust wav2vec 2.0: Analyzing Domain Shift in Self-Supervised Pre-Training.\vspace{1 mm}

[7] Yu Zhang, James Qin and Daniel S. Park et al. Pushing the Limits of Semi-Supervised Learning for Automatic Speech Recognition.\vspace{1 mm}

[8] J. Li, A. Sun, J. Han, and C. Li, ‘‘A survey on deep learning for named entity
recognition,’’ IEEE Trans. Knowl. Data Eng., vol. 34, no. 1, pp. 50–70,
Jan. 2022.\vspace{1 mm}

[9] G. Wu, G. Tang, Z. Wang, Z. Zhang and Z. Wang, "An Attention-Based BiLSTM-CRF Model for Chinese Clinic Named Entity Recognition," in IEEE Access, vol. 7, pp. 113942-113949, 2019, doi: 10.1109/ACCESS.2019.2935223.\vspace{1 mm}

[10] J. Diao, Z. Zhou and G. Shi, "Leveraging Integrated Learning for Open-Domain Chinese Named Entity Recognition," in International Journal of Crowd Science, vol. 6, no. 2, pp. 74-79, June 2022, doi: 10.26599/IJCS.2022.9100015.\vspace{1 mm}

[11] H. Čeović, A. S. Kurdija, G. Delač and M. Šilić, "Named Entity Recognition for Addresses: An Empirical Study," in IEEE Access, vol. 10, pp. 42108-42120, 2022, doi: 10.1109/ACCESS.2022.3167418.\vspace{1 mm}

[12] Juliette Millet, Charlotte Caucheteux and Pierre Orhan et al. Toward a realistic model of speech processing in the brain with self-supervised learning.\vspace{1 mm}

[13] W. Zaghloul and S. Trimi, ‘‘Developing an innovative entity extraction
method for unstructured data,’’ Int. J. Quality Innov., vol. 3, no. 1, pp. 1–10,
Dec. 2017, doi: 10.1186/s40887-017-0012-y.\vspace{1 mm}

[14] Alexei Baevski, Wei-Ning Hsu and Alexis Conneau et al. Unsupervised Speech Recognition. DOI:10.21437/Interspeech.2021-329\vspace{1 mm}

[15] Alexei Baevski, Henry Zhou, Abdelrahman Mohamed, and Michael Auli. 2020. Wav2vec 2.0: a framework for self-supervised learning of speech representations. In Proceedings of the 34th International Conference on Neural Information Processing Systems (NIPS'20). Curran Associates Inc., Red Hook, NY, USA, Article 1044, 12449–12460.

[16] Vaswani, Ashish, Shazeer, Noam, Parmar, Niki, Uszkoreit, Jakob, Jones, Llion, Gomez, Aidan N, Kaiser, Łukasz, and Polosukhin, Illia. Attention is all you need. In Advances in neural information processing systems, pp. 5998–6008, 2017.
\end{document}